\documentclass[sigconf]{acmart}

\usepackage{booktabs} 
\usepackage{subfigure}
\usepackage{color}
\usepackage{tabu}
\usepackage{physics}
\usepackage{epigraph}
\usepackage{ctable}
\usepackage{enumitem}
\usepackage{url}

\usepackage{algorithm,algpseudocode}
\usepackage{amsmath}
\usepackage{amsfonts}

\setlist[itemize]{leftmargin=*}

\copyrightyear{2022}
\acmYear{2022}
\setcopyright{rightsretained}
\acmConference[WWW '22]{Proceedings of the ACM Web Conference 2022}{April 25--29, 2022}{Virtual Event, Lyon, France}
\acmBooktitle{Proceedings of the ACM Web Conference 2022 (WWW '22), April 25--29, 2022, Virtual Event, Lyon, France}\acmDOI{10.1145/3485447.3512093}
\acmISBN{978-1-4503-9096-5/22/04}

\begin{document}

\title{MetaBalance: Improving Multi-Task Recommendations via Adapting Gradient Magnitudes of Auxiliary Tasks}


\author{Yun He}\thanks{*A majority of this work was done while the first author was interning at Meta AI}
\affiliation{%
 \institution{Texas A\&M University}
 \country{USA}
}
\email{yunhe@tamu.edu}
\author{Xue Feng}
\affiliation{%
 \institution{Meta AI}
  \country{USA}
}
\email{xfeng@fb.com}
\author{Cheng Cheng}
\affiliation{%
 \institution{Meta AI}
  \country{USA}
}
\email{cc6@fb.com}

\author{Geng Ji}
\affiliation{%
 \institution{Meta AI}
  \country{USA}
}
\email{gji@fb.com}

\author{Yunsong Guo}
\affiliation{%
 \institution{Meta AI}
  \country{USA}
}
\email{yunsong@fb.com}
\author{James Caverlee}
\affiliation{%
 \institution{Texas A\&M University}
  \country{USA}
}
\email{caverlee@tamu.edu}

\renewcommand{\shorttitle}{MetaBalance for Improving Multi-Task Recommendations}

\begin{abstract}
In many personalized recommendation scenarios, the generalization ability of a target task can be improved via learning with additional auxiliary tasks alongside this target task on a multi-task network. However, this method often suffers from a serious optimization imbalance problem. On the one hand, one or more auxiliary tasks might have a larger influence than the target task and even dominate the network weights, resulting in worse recommendation accuracy for the target task. On the other hand, the influence of one or more auxiliary tasks might be too weak to assist the target task. More challenging is that this imbalance dynamically changes throughout the training process and varies across the parts of the same network. We propose a new method: MetaBalance to balance auxiliary losses via directly manipulating their gradients w.r.t the shared parameters in the multi-task network. Specifically, in each training iteration and adaptively for each part of the network, the gradient of an auxiliary loss is carefully reduced or enlarged to have a closer magnitude to the gradient of the target loss, preventing auxiliary tasks from being so strong that dominate the target task or too weak to help the target task. Moreover, the proximity between the gradient magnitudes can be flexibly adjusted to adapt MetaBalance to different scenarios. The experiments show that our proposed method achieves a significant improvement of 8.34\% in terms of NDCG@10 upon the strongest baseline on two real-world datasets. The code of our approach can be found at here.\footnote{\url{https://github.com/facebookresearch/MetaBalance}}\end{abstract}

%
%
%




\maketitle

\section{Introduction}
\label{sec: introduction}

\begin{figure}[t]
  \centering
\subfigure[Transfer Learning from Auxiliary Tasks to Improve the Target Task on a Multi-task Network]{
    \label{fig: Transfer Learning from Auxiliary Tasks} 
    \includegraphics[width=3.12in]{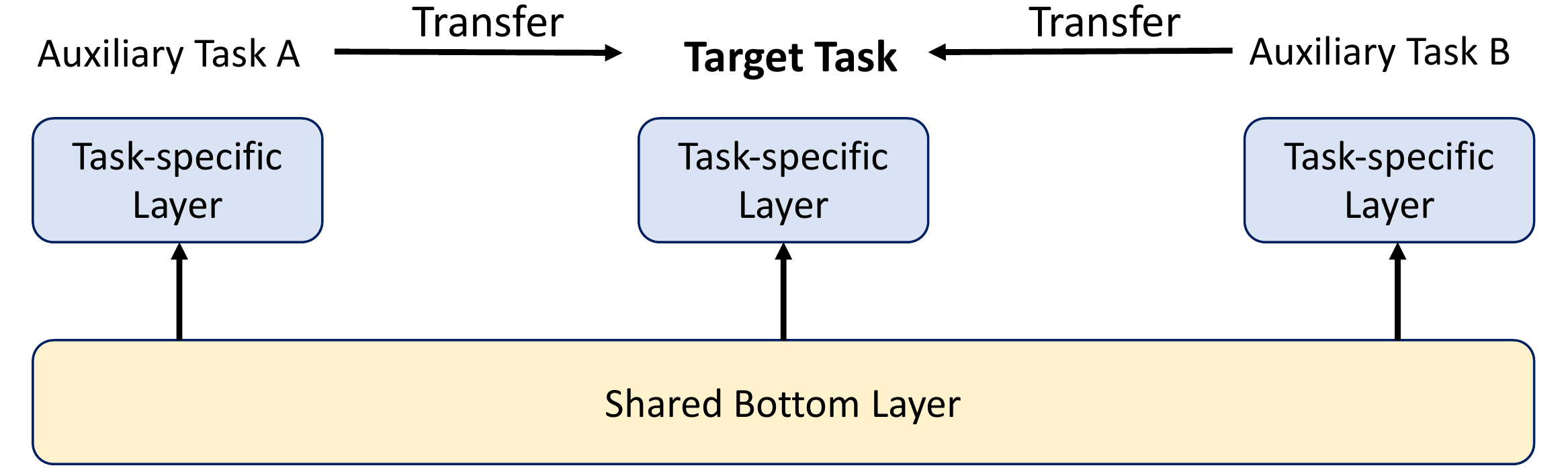}}

  \subfigure[Imbalance on one part of the shared parameters (e.g., MLP layer)]{
    \label{example: Imbalance on the MLP layer} 
    \includegraphics[width=1.62in]{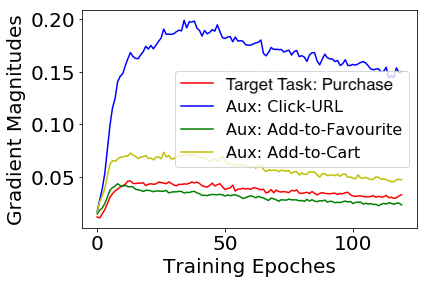}}
  \subfigure[Imbalance on another part of the shared parameters (e.g., embedding layer)]{
    \label{example: Imbalance on the user embedding layer} 
    \includegraphics[width=1.59in]{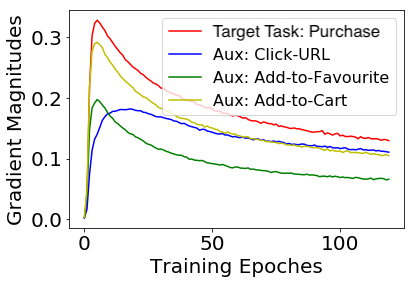}}
   \caption{The imbalance of gradient magnitudes in transfer learning from auxiliary tasks for recommendations on Alibaba data. The magnitudes dynamically change throughout the training, with the imbalance varying across different parts of the same multi-task network: in Fig \ref{example: Imbalance on the MLP layer}, the gradient of auxiliary task click-URL is much larger than the target gradient; in Fig \ref{example: Imbalance on the user embedding layer}, the gradient of auxiliary task Add-to-Favorite is much smaller than the target gradient. }
  \label{fig: examples of magnitude imbalance} 
\end{figure}

The accuracy of personalized recommendations can often be improved by transfer learning from related auxiliary information. For example, a primary task on e-commerce platforms like Amazon and eBay is to predict if a user will purchase an item. This purchase prediction task can benefit from transferring knowledge about the user's preference from auxiliary information like which item URLs the user has clicked and which items the user has put into the shopping cart. A common way to enable such transfer learning is to formulate this auxiliary information as auxiliary tasks (e.g., predict if a user will click a URL) and optimize them jointly with the target task (e.g., purchase prediction) on a multi-task network. In this way, knowledge can be transferred from the auxiliary tasks to the target task via the shared bottom layer of the multi-task network as shown in Figure \ref{fig: Transfer Learning from Auxiliary Tasks}. Enhanced with auxiliary information, the target task can obtain better performance than training the target task in isolation. Since the motivation of introducing those auxiliary tasks is often to purely assist the target task, in this paper, we focus on scenarios where only the performance of the target task is of interest. 

Beyond purchase prediction, many other recommendation scenarios \cite{ma2008sorec, guo2015trustsvd, wang2019social, ma2018entire, bansal2016ask, zhang2016collaborative, he2019hierarchical, cao2017embedding, liu2014recommending} can also benefit from such transfer learning from auxiliary tasks. In social recommendation \cite{ma2008sorec, guo2015trustsvd, wang2019social}, knowledge can be transferred from the social network to improve personalized recommendations  via training the target task simultaneously with auxiliary tasks like predicting the connections or trust among users. To better estimate post-click conversion rate (CVR) in online advertising, related information like post-view click-through rate (CTR) and post-view click-through \& conversion rate (CTCVR) can be introduced as auxiliary tasks \cite{ma2018entire}. Another example is that learning user and item embeddings from product review text can be designed as auxiliary tasks to improve the target goal of predicting ratings on e-commerce platforms \cite{zhang2016collaborative}.

However, a key challenge to transfer learning from auxiliary tasks in personalized recommendation is the potential for a significant \textit{imbalance of gradient magnitudes}, which can negatively affect the performance of the target task. As mentioned before, such transfer learning is often conducted on a multi-task network, which is commonly composed of a bottom layer with shared parameters and several task-specific layers. In training, each task has a corresponding loss and each loss has a corresponding gradient with respect to the shared parameters of that multi-task network. The sum of these gradients (for the target task and the auxiliary tasks) impacts how the shared parameters are updated. Hence, the larger the gradient is, the greater the impact this gradient has on the shared parameters. As a result, if the gradient of an auxiliary loss is much larger than the gradient of the target loss, the shared parameters will be most impacted by this auxiliary task rather than the target task. Consequently, the target task could be swamped by the auxiliary tasks, resulting in worse performance. On the other hand, if an auxiliary gradient is much smaller than the target gradient, the influence of this auxiliary task might be too weak to assist the target task. This \textit{imbalance of gradient magnitudes} is common in industrial recommender systems: Figure \ref{example: Imbalance on the MLP layer} and \ref{example: Imbalance on the user embedding layer} highlight two examples from Alibaba, which respectively demonstrate how the target task gradient can be dominated by an auxiliary task, and how some auxiliary tasks have gradients so small that they may only weakly inform the target task.

So how can we overcome this gradient imbalance? A simple and often used approach is to tune the weights of task losses (or gradients) through a grid or random search. However, such fixed task weights are not optimal because the gradient magnitudes change dynamically throughout the training and the imbalance might vary across the different subsets of the shared parameters as shown in Figure \ref{fig: examples of magnitude imbalance}. Besides, it is time-consuming to tune the weights for multiple auxiliary tasks.


In this paper, we propose MetaBalance as a novel algorithm and flexible framework that adapts auxiliary tasks to better assist the target task from the perspective of gradient magnitudes. Specifically, MetaBalance has three strategies: (A) Strengthening the dominance of the target task -- auxiliary gradients with larger magnitudes than the target gradient will be carefully reduced in each training iteration; (B) Enhancing the knowledge transferring from weak auxiliary tasks -- auxiliary gradients with smaller magnitudes than the target gradient will be carefully enlarged; and (C) MetaBalance adopts both (A) and (B) in the same iteration. In the absence of sufficient prior knowledge, which strategy to apply is treated as a data-driven problem, where the best strategy can be empirically selected based on the performance over the validation set of the target task. 

Moreover, MetaBalance has three key characteristics: 
\begin{enumerate}
	\item Auxiliary gradients can be balanced \textit{dynamically}  throughout the training process and \textit{adaptively} for different subsets of the shared parameters, which is more flexible than fixed weights for task losses;
	\item MetaBalance \textit{prioritizes} the target task via preventing auxiliary tasks from being so strong that they dominate the target task or too weak to help the target task, which can be easily monitored by choosing one of the three strategies;
	\item The next important question is \textit{how much should the auxiliary gradient magnitudes be reduced or enlarged?} We design a relax factor to control this to flexibly adapt MetaBalance to different scenarios. The relax factor can also be empirically selected based on the performance over the validation dataset of the target task.
\end{enumerate}

In sum, MetaBalance provides a flexible framework for adapting auxiliary gradients to better improve the target task from the perspective of gradient magnitudes. Extensive experiments over two real-world user behavior datasets from Alibaba show the effectiveness and flexibility of MetaBalance. In particular, we have four target observations:

\begin{itemize}
 \item With the best strategy and relax factor selected from the validation set, MetaBalance can significantly boost the test accuracy of the target task, which shows that auxiliary knowledge can be better transferred to the target task via MetaBalance.
 \item MetaBalance can significantly outperform previous methods for adapting auxiliary tasks to improve the target task. For example, we observe a significant improvement of 8.34\% upon the strongest baselines in terms of NDCG@10. 
 \item Only one hyper-parameter in MetaBalance (the relax factor) needs to be tuned, irrespective of the number of tasks. Hence, MetaBalance requires only a few training runs, which is more efficient than tuning the weights of task losses, which can be computationally intensive as the number of tasks increases.
 \item MetaBalance can collaborate well with several popular optimizers including Adam, Adagrad and RMSProp, which shows the potential that MetaBalance can be widely applied in many scenarios.
\end{itemize}

\section{Related Work}

\noindent\textbf{Recommendations with Auxiliary Tasks.} In many personalized recommendation scenarios, the test accuracy of the target task can be improved via joint learning with auxiliary tasks. In social recommendation \citep{ma2008sorec, guo2015trustsvd, wang2019social}, the knowledge about the user preference can be transferred from social network to the improve recommendations while the target task like rating prediction jointly train with auxiliary tasks like predicting the connections and trust among users. To improve post-click conversion rate (CVR) prediction, Ma et al. \citep{ma2018entire} consider the sequential pattern of user actions and introduce post-view click-through rate (CTR) and post-view click-through\&conversion rate (CTCVR) as auxiliary tasks. To enhance music playlists or booklists recommendations, predicting if a user will like an individual song or book can also be used as auxiliary tasks and jointly learn with the list-based recommendations. Besides, Bansal et al. \citep{bansal2016ask} design auxiliary tasks of predicting item meta-data (e.g., tags, genres) to improve the rating prediction as the target task. To improve the target goal of predicting ratings, learning user and item embeddings from product review text can also be designed as auxiliary tasks \cite{zhang2016collaborative}.

\smallskip
\noindent\textbf{Auxiliary Learning.} In this paper, we focus on transferring knowledge from auxiliary tasks to improve the target recommendation task, which is an example of \textit{auxiliary learning} paradigm. While multi-task learning aims to improve the performance across all tasks, auxiliary learning differs in that high test accuracy is only required for a primary task, and the role of the other tasks is to assist in generalization of the primary task. Auxiliary learning has been widely used in many areas. In speech recognition, Toshniwal et al. \citep{toshniwal2017multitask} apply auxiliary supervision from phoneme recognition to improve the performance of conversational speech recognition. In computer vision, Liebel at al. \citep{liebel2018auxiliary} propose auxiliary tasks such as the global description of a scene to boost the performance for single scene depth estimation. Mordan et al. \citep{mordan2018revisiting} observe that object detection can be enhanced if it jointly learns with depth prediction and surface normal prediction as auxiliary tasks. Liu et al. \citep{liu2019self} propose a Meta AuXiliary Learning (MAXL) framework that automatically learns appropriate labels for auxiliary tasks. In NLP, Trinh et al. \citep{trinh2018learning} show that unsupervised auxiliary losses significantly improve optimization and generalization of LSTMs. Auxiliary learning has also been applied to improve reinforcement learning \citep{jaderberg2016reinforcement, lin2019adaptive}.  

\smallskip
\noindent\textbf{Gradient Direction-based Methods for Adapting Auxiliary Tasks.} In auxiliary learning, several methods \citep{lin2019adaptive,yu2020gradient,du2018adapting} have been proposed to adapt auxiliary tasks to avoid the situation where they dominate or compete with the target task, where an auxiliary gradient will be down-weighted or masked out if its direction is conflicting with the direction of the target gradient. We will introduce these methods in detail in the Appendix (Section \ref{sec: Adapting Auxiliary Tasks Methods}) and compare them with the proposed MetaBalance. In particular, MetaBalance does not punish auxiliary gradients with conflict directions but strengths the dominance of the target task from the perspective of gradient magnitudes. In the experiments, MetaBalance shows better generalization than these gradient direction-based methods.




\medskip
\noindent\textbf{Multi-Task Learning.} Multi-task learning \citep{ruder2017overview, vandenhende2020multi} is used to improve the learning efficiency and prediction accuracy of multiple tasks via training them jointly. Shared-bottom model \citep{vandenhende2020multi} is a commonly used structure where task-specific tower networks receive the same representations that come from a shared bottom network. 


\medskip
\noindent\textbf{Multi-Task Balancing Methods.} In multi-task learning, methods have been proposed to balance the joint learning of all tasks to avoid the situation where one or more tasks have a dominant influence on the network weight \citep{kendall2018multi, chen2018gradnorm, malkiel2020mtadam, liu2019end, sener2018multi}. Although these methods have no special preference to the target task (as in our focus in this paper), we do discuss their connection to MetaBalance in Section \ref{sec:Multi-Task Balancing Methods} (in Appendix) and experimentally compare with them in Section ~\ref{sec:experiments}.  

\section{Problem Statement}
\label{sec: Problem Formulation}
Our goal is to improve the test accuracy of a target task via training auxiliary tasks alongside this target task on a multi-task network, where useful knowledge from auxiliary tasks can be transferred so that the shared parameters of the network converge to more robust features for the target task. In the context of personalized recommendation, the target task is normally to predict if a user will interact (e.g., purchase or click) with an item, which can be formulated as a binary classification problem. The test accuracy is measured over the top-K items ranked by their probabilities of being interacted with by the user against the ground-truth set of items that the user actually interacted with.

Let $\theta$ denote a subset of the shared parameters. For example, $\theta$ could be the weight matrix or the bias vector of a multi-layer perceptron in the shared bottom network. $\theta$ is learned by jointly minimizing the target task loss $\mathcal{L}_{tar}$ with auxiliary task losses $\mathcal{L}_{aux,i}, i=1,...,K$:
\begin{equation}
	\mathcal{L}_{total} = \mathcal{L}_{tar} + \sum_{i=1}^{K} \mathcal{L}_{aux,i}
\end{equation}

We assume that we update $\theta^{t}$ via gradient descent with learning rate $\alpha$:
\begin{equation}
	\label{equation:update}
	\theta^{t+1} = \theta^{t} -\alpha * \mathbf{G}_{total}^{t}
\end{equation}
where $t$ means the $t$-th training iteration over the mini-batches ($t=1, 2...T$) and  $\mathbf{G}_{total}^{t}$ is the gradient of $\mathcal{L}^{t}_{total}$ w.r.t $\theta$:
\begin{gather}
	\label{equation:gradient sum}
	\mathbf{G}_{total}^{t} = \nabla_{\theta} \mathcal{L}^{t}_{total} =  \nabla_{\theta} \mathcal{L}^{t}_{tar} + \sum_{i=1}^{K}  \nabla_{\theta} \mathcal{L}_{aux,i}^{t} 
\end{gather}
where $\mathbf{G}_{total}$ is equivalent to adding up each gradient of the target and auxiliary losses. To simplify the notations, we have:

\begin{itemize}
	\item $\mathbf{G}_{tar}$ (i.e., $\nabla_{\theta} \mathcal{L}_{tar}$): the gradient of the target task loss $\mathcal{L}_{tar}$ with respect to $\theta$. 
	\item $\mathbf{G}_{aux,i}$ (i.e., $\nabla_{\theta} \mathcal{L}_{aux,i}$): the gradient of the $i$-th auxiliary task loss $\mathcal{L}_{aux,i}$ with respect to $\theta$, where $i=1, 2...K$.
	\item $\| \mathbf{G} \|$: the magnitude (L2 Norm) of the corresponding gradient.
\end{itemize}

As shown in Eq \ref{equation:gradient sum} and \ref{equation:update}, the larger the magnitude of a gradient is, the greater the influence this gradient has in updating $\theta$. 

%

\section{Proposed Method}
The imbalance of gradient magnitudes may negatively affect the target task optimization. On the one hand, if $\| \mathbf{G}_{aux,i} \|$ ($\exists i\in\{1, 2...K\}$) is much larger than $\| \mathbf{G}_{tar} \|$, the target task will lose its dominance of updating $\theta$ and get lower performance. On the other hand, if $\| \mathbf{G}_{aux,i} \|$ ($\exists i\in\{1, 2...K\}$) is much smaller than $\| \mathbf{G}_{tar} \|$, the corresponding auxiliary task might become less influential to assist the target task. As illustrated in Figure \ref{fig: examples of magnitude imbalance}, many personalized recommendations may suffer from this imbalance.  Hence, we are motivated to propose a new algorithm that adapts auxiliary tasks from the perspective of gradient magnitudes. 


\begin{algorithm}[H] 
\caption{The Basic Version of MetaBalance}
\small
\label{alg:Basic MetaBalance}
\begin{algorithmic}[1]
\Require{$\theta^{1}$, $\mathcal{L}_{tar}$, $\mathcal{L}_{aux,1},...,\mathcal{L}_{aux,K}$, \textit{\textbf{Strategy}} that is selected from \{$\| \mathbf{G}_{aux,i} \|> \| \mathbf{G}_{tar} \|$, $\| \mathbf{G}_{aux,i} \|< \| \mathbf{G}_{tar} \|$, ($\| \mathbf{G}_{aux,i} \|> \| \mathbf{G}_{tar} \|$) or ($\| \mathbf{G}_{aux,i} \|< \| \mathbf{G}_{tar} \|$)\}}
\Ensure{$\theta^{T}$}
\Statex

\For{t = 1 to T}
	\State {$\mathbf{G}_{tar}^{t} = \nabla_{\theta} \mathcal{L}_{tar}^{t}$}
\For{i = 1 to K}  
	\State {$\mathbf{G}_{aux,i}^{t} = \nabla_{\theta} \mathcal{L}_{aux,i}^{t}$}   
	\If{(\textit{Strategy})}
    \State {$\mathbf{G}_{aux,i}^{t} \leftarrow \mathbf{G}_{aux,i}^{t} * \dfrac{\| \mathbf{G}_{tar}^{t} \|}{\| \mathbf{G}_{aux,i}^{t} \|} $ } 
    \EndIf
\EndFor
\State {$\mathbf{G}_{total}^{t} = \mathbf{G}_{tar}^{t} + \mathbf{G}_{aux,1}^{t} +  \dots \mathbf{G}_{aux, K}^{t}$} (element-wise addition)
\State Update $\theta$ using $\mathbf{G}_{total}^{t}$ (e.g., Gradient Descent: $\theta^{t+1} = \theta^{t} -\alpha * \mathbf{G}_{total}^{t}$)
\EndFor

\end{algorithmic}
\end{algorithm}

\subsection{Adapting Auxiliary Gradient Magnitudes}
As discussed above, the magnitude imbalance between $\mathbf{G}_{tar}$ and $\mathbf{G}_{aux,i},...,\mathbf{G}_{aux, K}$ may negatively affect the target task optimization. To alleviate this imbalance, MetaBalance is proposed to dynamically and adaptively balance the magnitudes of auxiliary gradients with three strategies and a relax factor (will be detailed in the next subsection). 

The basic version of MetaBalance is presented in Algorithm \ref{alg:Basic MetaBalance}, including four steps:
\begin{enumerate}[leftmargin=0cm,itemindent=.5cm,labelwidth=\itemindent,labelsep=0cm,align=left] 
	\item \textbf{Calculating the Gradients.} In each training iteration, we firstly calculate $\mathbf{G}_{tar}^{t}$ and $\mathbf{G}_{aux,i}^{t}$ respectively (line 2 and 4). 
	\item \textbf{Applying the Strategy.} In line 5, we can choose either reducing auxiliary gradients with larger magnitudes than the target gradient, or enlarging auxiliary gradients with smaller magnitudes, or applying the two strategies together. The strategy can be selected based on the validation performance of the target task. 
	\item \textbf{Balancing the Gradients.} Next, $\mathbf{G}_{aux,i}^{t}$ is normalized to be a unit vector by dividing by $\| \mathbf{G}_{aux,i}^{t} \|$ and then rescaled to have the same magnitude as $\mathbf{G}_{tar}^{t}$ by multiplying $\| \mathbf{G}_{tar}^{t} \|$ (line 6). 
	\item \textbf{Updating the Parameters.} After that, $\mathbf{G}_{total}^{t}$ (line 9) is obtained by summing $\mathbf{G}_{tar}^{t}$ and balanced $\mathbf{G}_{aux, 1}^{t}, \dots \mathbf{G}_{aux, K}^{t}$ together. Then, $\mathbf{G}_{total}^{t}$ is used to update $\theta$ following an optimizer's rule such as gradient descent (line 10). Since step (3) and (4) are completely decoupled, MetaBalance has the potential to collaborate with most commonly used optimizers like Adam and Adagrad \citep{duchi2011adaptive}.
\end{enumerate}

MetaBalance benefits auxiliary learning from six aspects: 
\begin{enumerate}[leftmargin=0cm,itemindent=.5cm,labelwidth=\itemindent,labelsep=0cm,align=left] 
	\item $ \mathbf{G}^{t}_{aux,i} $ with much larger magnitude than $ \mathbf{G}^{t}_{tar} $ could be automatically reduced, which prevents the dominance of one or more auxiliary tasks for the target task. (Strategy A)
	\item $ \mathbf{G}^{t}_{aux,i} $ with much smaller magnitude than $ \mathbf{G}^{t}_{tar} $ could be automatically enlarged, which enhances the knowledge transference from the corresponding auxiliary task. (Strategy B) 
	\item The (1) and (2) could be done together if necessary. (Strategy C)
	\item The strategy is selected based on the target task's performance over validation dataset, which is the empirically best strategy for a specific task and dataset. 
	\item Because $\frac{\| \mathbf{G}_{tar}^{t} \|}{\| \mathbf{G}_{aux,i}^{t} \|}$ can be regarded as a dynamic weight for $\mathbf{G}_{aux,i}^{t}$ in line 6, MetaBalance can balance $ \mathbf{G}_{aux,i}^{t} $ dynamically throughout the training process.
	\item As shown in Figure \ref{fig: examples of magnitude imbalance}, the imbalance of gradient magnitudes varies across the different parts of the same network (e.g., the auxiliary gradients might be much larger than the target gradient in an MLP but much smaller in an embedding layer).    Because MetaBalance can be easily applied to each part of the shared parameters separately ($\theta$ is an input of Algorithm \ref{alg:Basic MetaBalance}), the training of the different parts can be balanced respectively and adaptively. (5) and (6) makes MetaBalance more flexible than using fixed weights for task losses.
\end{enumerate}


\begin{figure*}
  \centering
\setlength{\abovecaptionskip}{0.0cm}
  \setlength{\belowcaptionskip}{-0.2cm}
 \subfigure[MetaBalance with $r=0$]{
    \label{visual: MetaBalance-Fix with r=0}
    \includegraphics[width=1.6in]{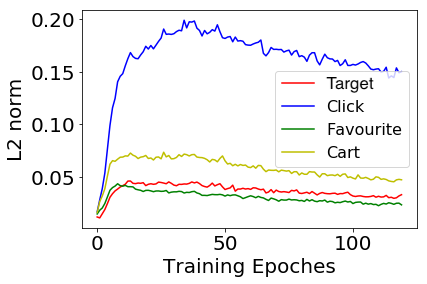}}
 \subfigure[MetaBalance with $r=0.2$]{
    \label{visual: MetaBalance-Fix with r=0.2} 
    \includegraphics[width=1.6in]{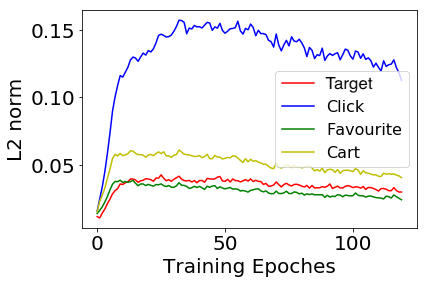}} 
 \subfigure[MetaBalance with $r=0.7$]{
    \label{visual: MetaBalance-Fix with r=0.7} 
    \includegraphics[width=1.6in]{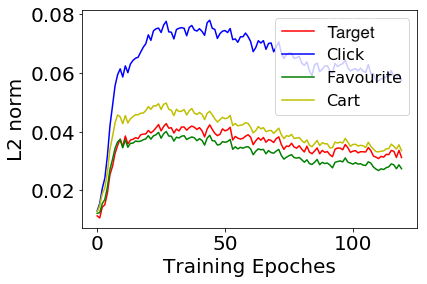}}   
 \subfigure[MetaBalance with $r=1.0$]{
    \label{visual: MetaBalance-Fix with r=1} 
    \includegraphics[width=1.6in]{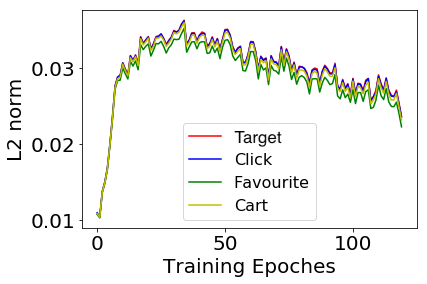}}
   \caption{The impact of relax factor $r$ on magnitude proximity on UserBehavior-2017 dataset. In the legend, ``target'' represents the target task (i.e., purchase prediction). Y-axis is the average gradient magnitude over all mini-batch iterations in one epoch, where the gradient w.r.t a MLP layer of the multi-task network is taken as the example. }
  \label{fig: visualization of metabalance} 
\end{figure*}

However, the drawback of this basic version in Algorithm \ref{alg:Basic MetaBalance} is also obvious: forcing auxiliary gradients to have exactly the same magnitude as the target gradient might not be optimal for the target task. To overcome this inflexibility of the magnitude scaling, we design a relax factor to control the closeness of $\| \mathbf{G}_{aux,i}^{t} \|$ to $\| \mathbf{G}_{tar}^{t} \|$ in the following subsection.

\subsection{Adjusting Magnitude Proximity}
\label{sec: Relaxation Function}
The next question is how to flexibly adjust the magnitude proximity between $\mathbf{G}_{aux,i}$ and $\mathbf{G}_{tar}$ to adapt to different scenarios? We design a relax factor $r$ to control this magnitude proximity, which is used in line 6 of Algorithm \ref{alg:Basic MetaBalance}:
$$
\mathbf{G}_{aux,i}^{t} \leftarrow (\mathbf{G}_{aux,i}^{t} * \dfrac{\| \mathbf{G}_{tar}^{t} \|}{\| \mathbf{G}_{aux,i}^{t} \|} )*  r + \mathbf{G}_{aux,i}^{t} * (1-r)
$$
where, if $r=1$, then $\mathbf{G}_{aux,i}^{t}$ has exactly the same magnitude as $\mathbf{G}_{tar}^{t}$. If $r=0$, then $\mathbf{G}_{aux,i}^{t}$ keeps its original magnitude. The larger $r$ is, the closer $\| \mathbf{G}_{aux,i}^{t} \|$ gets to $\|\mathbf{G}_{tar}^{t} \|$. Hence, $r$ balances the magnitude information between each auxiliary gradient and the target gradient. 

The impact of $r$ on the magnitude proximity is illustrated in Figure \ref{fig: visualization of metabalance}. We observe that the target gradient is dominated by an auxiliary gradient with its much larger magnitude when $r=0$ in Figure \ref{example: Imbalance on the MLP layer}. In contrast, $r=1$ lets all gradients have the same but very small magnitude as the target gradient in Figure \ref{visual: MetaBalance-Fix with r=1}. Between the two extremes, Figure \ref{visual: MetaBalance-Fix with r=0.2} ($r=0.2$) and Figure \ref{visual: MetaBalance-Fix with r=0.7} ($r=0.7$) balance the gradient magnitudes in a more moderate way, which pushes $\| \mathbf{G}_{aux,i}^{t} \|$ closer to $\| \mathbf{G}_{tar}^{t} \|$ but not exactly the same -- the original magnitude can be partially kept.


More than that, \textit{$r$ can actually affect the weight for each auxiliary task.} We can further reformulate line 6 in Algorithm \ref{alg:Basic MetaBalance} as:
$$
	\mathbf{G}_{aux,i}^{t} \leftarrow \mathbf{G}_{aux,i}^{t} * w_{aux,i}^{t} \\
$$ where $w_{aux,i}^{t}$ is the weight for $\mathbf{G}_{aux,i}^{t}$ and we have:
\begin{gather}
\label{equa:r affects weight of auxiliary task}
w_{aux,i}^{t} = (\dfrac{\| \mathbf{G}_{tar}^{t} \|}{\| \mathbf{G}_{aux,i}^{t} \|} -1)*  r + 1
\end{gather}
where, if $\| \mathbf{G}_{tar}^{t} \| >\| \mathbf{G}_{aux,i}^{t} \| $, the higher $r$ is, the higher $w_{aux,i}^{t}$ will be; however, if $\| \mathbf{G}_{tar}^{t} \| < \| \mathbf{G}_{aux,i}^{t} \| $, the higher $r$ is, the lower  $w_{aux,i}^{t}$ will be. 

The next key question is how to choose this $r$? As presented in Equation \ref{equa:r affects weight of auxiliary task}, $r$ affects the weight for each auxiliary task. Without the prior knowledge of the importance of each auxiliary task to the target task, we treat the setting of $r$ as a data-driven problem and believe that $r$ should be carefully adjusted to adapt to different scenarios. Since $r$ is only used in the backward propagation and hence has no gradient from any loss, $r$ is not a learnable parameter inherently. Hence, we treat $r$ as a hyper-parameter, which is tuned over validation datasets. Note that the same $r$ for all auxiliary tasks does not mean that they will have the same weight or gradient magnitude because $w_{aux,i}^{t}$ is not only decided by $r$ but also affected by $\| \mathbf{G}_{tar}^{t} \|$ and $\| \mathbf{G}_{aux,i}^{t} \|$ (see Equation \ref{equa:r affects weight of auxiliary task}). 

Therefore, there is only one hyper-parameter $r$ in MetaBalance that needs to be tuned, which is irrespective of the number of tasks. In contrast, the computational complexity of tuning weights of task losses increases exponentially for each task added. Moreover, we also observe that MetaBalance achieves higher test accuracy than tuning the task weights in our experiments.

Finally, instead of using current magnitudes $\| \mathbf{G}_{tar}^{t} \|$ and $\| \mathbf{G}_{aux,i}^{t} \|$ in Algorithm \ref{alg:Basic MetaBalance}, following \citep{malkiel2020mtadam}, we apply the moving average of magnitude of the corresponding gradient to take into account the variance among all gradient magnitudes over the training iterations:
\begin{gather}
\label{equation: moving average}
	m_{tar}^{t} = \beta*m_{tar}^{t-1} + (1-\beta)*\| \mathbf{G}_{tar}^{t} \| \\
\label{equation: moving average auxiliary task}
	m_{aux,i}^{t} = \beta*m_{aux,i}^{t-1} + (1-\beta)*\| \mathbf{G}_{aux,i}^{t} \| , \forall i=1,...,K
\end{gather}
where $m_{tar}^{0} = m_{aux,i}^{0} = 0$ and $\beta$ is to control the exponential decay rates of the moving averages, which could be empirically set as 0.9. The moving averages make the training more stable and will be discussed in the experiments. Finally, the complete version of MetaBalance is shown in Algorithm \ref{alg:The Complete Version of MetaBalance}. 


\subsection{Time and Space Complexity Analysis} 
\label{sec: Time and Space Complexity Analysis of MetaBalance}
In this section, we show that MetaBalance does \textbf{not} significantly increase the time and space complexity of training multi-task networks. Assume that addition, subtraction, multiplication, division and square root take ``one unit'' of time. The time complexity of training a multi-task network depends on the network's structure. For simplicity, assume that an MLP is the shared layer of a multi-task network and $\theta$ is a weight matrix of a single layer in the MLP, where $\theta$ has input dimension $n$ and output dimension $m$. The time complexity of updating $\theta$ is $\mathcal{O}(T(1+K)nmd)$ \citep{baur1983complexity}, where $T$ is the number of training iterations over the mini-batches, $(1+K)$ is the count of the target task plus $K$ auxiliary tasks and $d$ is the size of the mini-batch. For MetaBalance in Algorithm \ref{alg:The Complete Version of MetaBalance}, in each training iteration and for each task, $r$ is a hyper-parameter, calculating $m_{aux,i}^{t}$ or $m_{tar}^{t}$ takes $\mathcal{O}(nmd)$, and the time complexity of updating the magnitude of $\mathbf{G}_{aux,i}^{t}$ (line 9) is also $\mathcal{O}(nmd)$. To sum up, the time complexity of MetaBalance is still $\mathcal{O}(T(1+K)nmd)$. Therefore, MetaBalance will not significantly slow down the training of multi-task networks. 

\begin{algorithm}[H] 
\caption{The Complete Version of MetaBalance}
\small
\label{alg:The Complete Version of MetaBalance}
\begin{algorithmic}[1]
\Require{$\theta^{1}$, $\mathcal{L}_{tar}$, $\mathcal{L}_{aux,1},...,\mathcal{L}_{aux,K}$, relax factor $r$, $\beta$ in moving averages, \textit{\textbf{Strategy}} that is selected from \{$ m_{aux,i} >  m_{tar} $, $ m_{aux,i} <  m_{tar} $, ($ m_{aux,i} >  m_{tar} $) or ($ m_{aux,i} <  m_{tar} $)\}} 
\Ensure{$\theta^{T}$}
\Statex


\State {Initialize $m_{tar}^{0} = m_{aux,i}^{0}=0$}


\For{t = 1 to T}
	
	\State {$\mathbf{G}_{tar}^{t} = \nabla_{\theta} \mathcal{L}_{tar}^{t}$}
	\State {$m_{tar}^{t} = \beta*m_{tar}^{t-1} + (1-\beta)*\| \mathbf{G}_{tar}^{t} \|$}
\For{i = 1 to K}  
	\State {$\mathbf{G}_{aux,i}^{t} = \nabla_{\theta} \mathcal{L}_{aux,i}^{t}$}   
	\State {$m_{aux,i}^{t} = \beta*m_{aux,i}^{t-1} + (1-\beta)*\| \mathbf{G}_{aux,i}^{t} \|$}
	\If{(\textit{Strategy})}
    \State {$\mathbf{G}_{aux,i}^{t} \leftarrow (\mathbf{G}_{aux,i}^{t} * \dfrac{m_{tar}^{t}}{m_{aux,i}^{t}} )*  r + \mathbf{G}_{aux,i}^{t} * (1-r)$ } 
    \EndIf
\EndFor
\State {$\mathbf{G}_{total}^{t} = \mathbf{G}_{tar}^{t} + \mathbf{G}_{aux,1}^{t} +  \dots \mathbf{G}_{aux,K}^{t}$} (element-wise addition)
\State Update $\theta$ using $\mathbf{G}_{total}^{t}$ (e.g., $\theta^{t+1} = \theta^{t} -\alpha * \mathbf{G}_{total}^{t}$)
\EndFor

\end{algorithmic}
\end{algorithm}

Except for the space of training a multi-task network, MetaBalance only requires extra space for $m_{tar}$, $r$, $\beta$ and $m_{aux,i}$,..., $m_{aux, K}$, where the space complexity is $\mathcal{O}(3+K)=\mathcal{O}(1)$ ($K$ is normally a small number). Hence, MetaBalance does not significantly increase the space complexity of multi-task networks training either.

\section{Experiments}
\label{sec:experiments}
In this section, we present our results and discussion toward answering the following experimental research questions:
\begin{itemize}
	 \item \textbf{RQ1:} How well does MetaBalance improve the target task via adapting the magnitudes of auxiliary gradients?
	\item \textbf{RQ2:} How well does MetaBalance perform compared to previous auxiliary task adapting and multi-task balancing methods?
	\item \textbf{RQ3:} How well does MetaBalance collaborate with commonly used optimizers such as Adam and Adagrad?
	\item \textbf{RQ4:} What is the impact of moving averages of gradient magnitudes in MetaBalance?
\end{itemize}

\subsection{Experimental Setup} 
Following the auxiliary learning setting \citep{valada2018deep,liu2019self}, high test accuracy is only required for a target task while the role of auxiliary tasks is to assist the target task to achieve better test accuracy. 

\medskip
\noindent\textbf{Datasets.} IJCAI-2015\footnote{\url{https://tianchi.aliyun.com/dataset/dataDetail?dataId=47&userId=1}} is a public dataset from IJCAI-15 contest, which contains millions of anonymized users' shopping logs in the past 6 months. UserBehavior-2017\footnote{\url{https://tianchi.aliyun.com/dataset/dataDetail?dataId=649&userId=1}} is a public dataset of anonymized user behaviors from Alibaba. The two datasets both contain users' behaviors including click, add-to-cart, purchase and add-to-favorite. The statistics of preprocessed datasets are summarized in Table \ref{tab:Statistics of datasets} (in Appendix). We treat purchase prediction as the target task and the prediction of other behaviors as auxiliary tasks. We formulate the prediction of each behavior like purchase as a binary classification problem and negative samples are randomly selected.

\medskip
\noindent\textbf{Evaluation and metrics.} In the evaluation, \textbf{all items} are ranked according to the probability of being purchased by the user and the top-K items are returned and measured against the ground-truth items set of what users actually purchased, where we adopt three metrics: Normalized Discounted Cumulative Gain (NDCG) \citep{jarvelin2002cumulated} at 10 and 20 (N@10 and N@20), precision at 10 and 20 (P@10 and P@20), and recall at 10 and 20 (R@10 and R@20). 

\medskip
\noindent\textbf{Multi-task network.} Because how to design a better multi-task network is not the emphasis of this paper, we directly adopt the combination of MLP layer and matrix factorization layer as the shared bottom network, which is widely adopted for recommendations in both academia \citep{ncf} and industry like Google \cite{cheng2016wide} and Facebook \cite{naumov2019deep}. We build MLP layer as the task-specific tower for each task. The multi-task network is shown in Figure \ref{fig: Multi-Task Recommendation Network} (Appendix). 

\medskip
\noindent\textbf{Baselines.} We compare MetaBalance with 10 baseline methods. Gradient direction-based methods that are designed for adapting auxiliary tasks to improve the target task, which will be detailed in Section \ref{sec: Adapting Auxiliary Tasks Methods}, including: \textbf{GradSimilarity} \citep{du2018adapting}, \textbf{GradSurgery} \citep{yu2020gradient},  \textbf{OL-AUX} \citep{lin2019adaptive}. Multi-Task balancing methods that treat all tasks equally, which will be detailed in Section \ref{sec:Multi-Task Balancing Methods}, including: \textbf{Uncertainty} \citep{kendall2018multi}, \textbf{GradNorm} \citep{chen2018gradnorm}, \textbf{DWA} \citep{liu2019end}, \textbf{MTAdam} \citep{malkiel2020mtadam} and \textbf{MGDA} \citep{sener2018multi}. And three simple baselines. \textbf{Single-Loss}: we mask out the loss terms of auxiliary tasks and only use target task loss to calculate gradients and update parameters in the model. \textbf{Vanilla-Multi}: multiple loss terms are not balanced where the weights for all loss terms are 1. \textbf{Weights-Tuning}: weights of loss terms are obtained by random search.

\medskip
\noindent\textbf{Reproducibility.} Due to limited space, the details of reproducibility is presented in Appendix (Section \ref{appen: Reproducibility}), including dataset preprocessing and split, implementation and training details.




\subsection{RQ1: Improvement of Target Task via Adapting Auxiliary Gradients}
\label{sec: Impact of Relax Factor}
In this subsection, we discuss the impact of adapting auxiliary gradient magnitudes on the target task's performance.

\begin{table*}[htbp]
  \centering
\small
    \setlength{\abovecaptionskip}{0.0cm}
  \setlength{\belowcaptionskip}{0.85cm}
\renewcommand\arraystretch{1.0}
  \caption{Strategy Selection}
    \begin{tabular}{cccc|ccc}
\toprule
    Datasets      & \multicolumn{3}{c|}{UserBehavior-2017} & \multicolumn{3}{c}{IJCAI-2015} \\
\midrule
    Metrics (\%)     & N@10  & R@10  & P@10  & N@10  & R@10  & P@10 \\
\midrule
    Vanilla-Multi & 0.820 & 1.284 & 0.291  & 0.844 & 0.965 & 0.437  \\
    Strategy A (strengthening the dominance of the target task)  & 0.948 & 1.487 & 0.316 & 0.858 & 0.963 & 0.424\\
    Strategy B (enhancing the knowledge transferring from weak auxiliary tasks)  & 0.904 & 1.384 & 0.301 & 0.818 & 0.950 & 0.425 \\
    Strategy C (Adopting Strategy A and Strategy B together) & 0.990 & 1.550 & 0.339 & 0.974 & 1.164 & 0.509 \\
\bottomrule
    \end{tabular}%
  \label{tab:Only Reducing or Enlarging Auxiliary Gradients Respectively}%
\end{table*}%

\medskip
\noindent\textbf{Impact of Strategy Selection.} We firstly study which strategy is optimal for the two recommendation datasets.  Note that we firstly compare the three strategies over the validation dataset to choose the best one and apply it on the test dataset. To be consistent with other experimental results, we present the results of the three strategies over the test dataset in Table \ref{tab:Only Reducing or Enlarging Auxiliary Gradients Respectively}, which reflects the same pattern as the validation dataset. First of all, all three strategies significantly outperform vanilla multi-task learning baseline (``Vanilla-Multi") in UserBehavior-2017 and Strategy C significantly outperforms the baseline in IJCAI-2015, which shows the effectiveness and robustness of MetaBalance. We observe the pattern ``Strategy C > Strategy A > Strategy B" across the two datasets, which shows that strengthening the dominance of the target task (Strategy A) is more important than enhancing the knowledge transferring from weak auxiliary tasks (Strategy B) and combining the two strategies together can achieve further improvements for the two datasets. Therefore, we apply Strategy C in the rest of the experiments.

\begin{figure}[]
    \centering
    \setlength{\abovecaptionskip}{0.0cm}
  \setlength{\belowcaptionskip}{-0.5cm}
    \includegraphics[width=.65\linewidth]{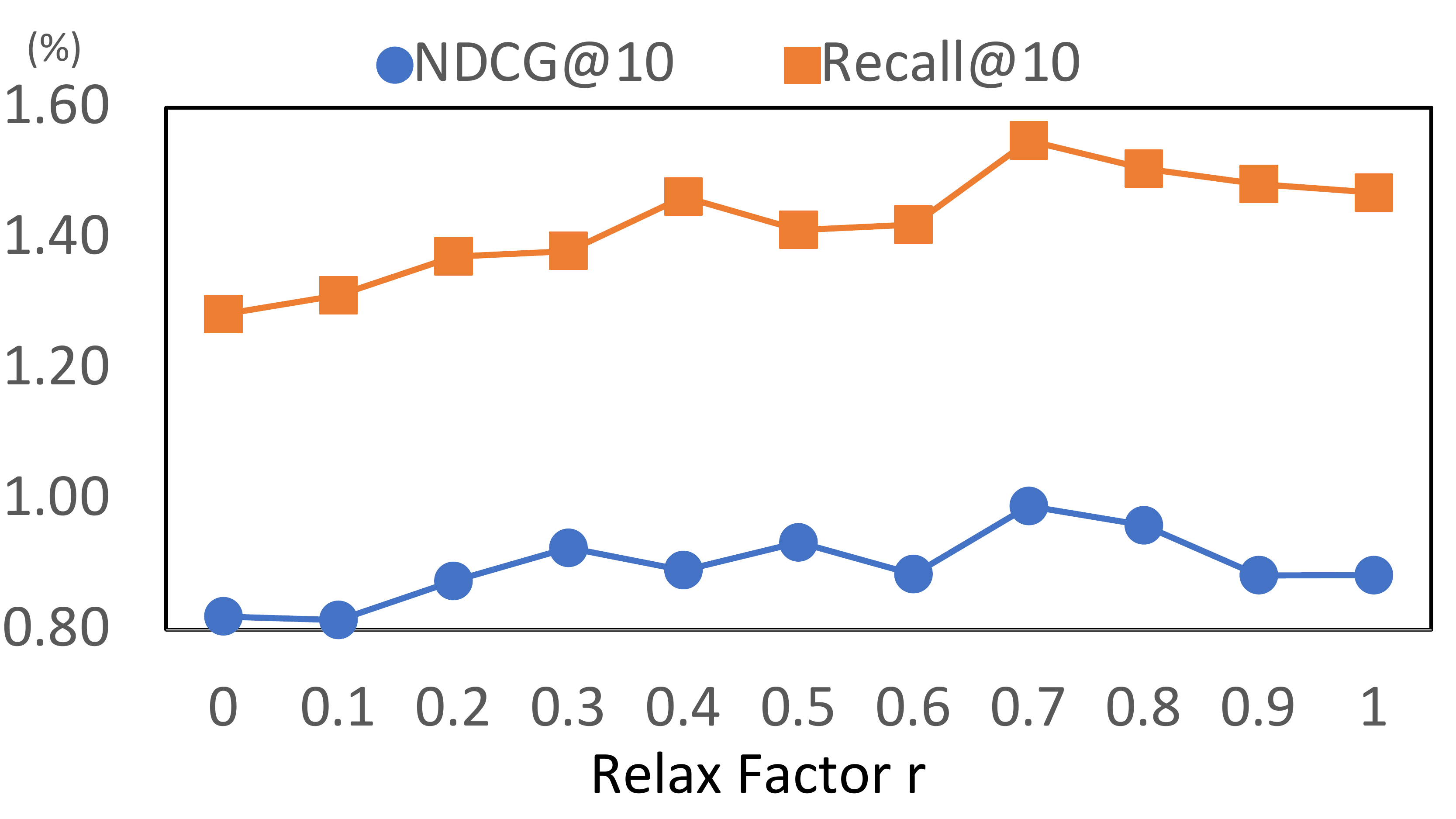}
    \caption{Impact of relax factor $r$}
    \label{impact of relax factor on public datasets}
\end{figure}

\medskip
\noindent\textbf{Impact of Relax Factor.}  Based on Strategy C, we further study the impact of the relax factor. Figure \ref{impact of relax factor on public datasets} presents the varying of NDCG@10 and Recall@10 as $r$ changes in UserBehavior-2017 dataset (the similar observation is obtained in IJCAI-2015 dataset).

The worst NDCG@10 and Recall@10 are achieved when $r=0$ (Vanilla-Multi), where auxiliary gradients ($\| \mathbf{G}_{aux,i}^{t} \|$) keep their original magnitudes (i.e., not balanced as in Vanilla-Multi). In Figure \ref{example: Imbalance on the MLP layer}, we observe that click task's gradient magnitude (blue curve) is much larger than $\| \mathbf{G}_{tar}^{t} \|$ (red curve). Hence, the target task gradient is probably dominated by the click task gradient, which explains the low target task performance of Vanilla-Multi (see Table \ref{tab:Experimental Results}).

In contrast, $r=1$ in MetaBalance means that $\| \mathbf{G}_{aux,i}^{t} \|$ becomes very close to $\| \mathbf{G}_{tar}^{t} \|$ as shown in Figure \ref{visual: MetaBalance-Fix with r=1}, where the four curves are twisted together. However, $r=1$ achieves suboptimal results as shown in Figure \ref{impact of relax factor on public datasets}, which demonstrates that the target task performance might be negatively impacted by a large $r$. A possible reason is that most auxiliary gradients are reduced to be very similar to the target gradient and hence the update of the shared parameters becomes so small that it negatively affects the optimization.

Between the two extremes, Figure \ref{visual: MetaBalance-Fix with r=0.2} ($r=0.2$) and Figure \ref{visual: MetaBalance-Fix with r=0.7} ($r=0.7$) balance the gradient magnitudes in a more moderate way -- getting $\| \mathbf{G}_{aux,i}^{t} \|$ closer to $\| \mathbf{G}_{tar}^{t} \|$ but not exactly the same, where $r=0.7$ achieves the best performance as shown in Figure \ref{impact of relax factor on public datasets}.

\begin{table*}[htbp]
  \centering
   \setlength{\abovecaptionskip}{0.0cm}
  \setlength{\belowcaptionskip}{0.0cm}
\small
\renewcommand\arraystretch{0.85}
\setlength{\tabcolsep}{6.5pt}
  \caption{Experimental Results}
    \begin{tabular}{l|cccccc|cccccc}
	\toprule
          & \multicolumn{6}{c|}{UserBehavior-2017}         & \multicolumn{6}{c}{IJCAI-2015} \\
	\midrule
    Metric(\%) & N@10  & R@10  & P@10  & N@20  & R@20  & P@20  & N@10  & R@10  & P@10  & N@20  & R@20  & P@20 \\
	\midrule
    Single-Loss & 0.817 & 1.265 & 0.275 & 0.994 & 1.825 & 0.208 & 0.883 & 0.935 & 0.431 & 1.022 & 1.314 & 0.298 \\
    Vanilla-Multi & 0.820 & 1.284 & 0.291 & 1.074 & 2.107 & 0.237 & 0.844 & 0.965 & 0.437 & 0.992 & 1.353 & 0.311 \\
    Weights-Tuning & 0.909 & 1.378 & \underline{0.326} & 1.165 & 2.195 & 0.263 & 0.866 & 1.013 & 0.445 & 1.037 & 1.448 & 0.330 \\
	\midrule
    Uncertainty & 0.724 & 1.158 & 0.266 & 0.903 & 1.739 & 0.201 & 0.695 & 0.818 & 0.365 & 0.834 & 1.186 & 0.266 \\
    GradNorm & 0.913 & 1.292 & 0.297 & 1.147 & 2.044 & 0.237 & 0.878 & 0.953 & 0.430 & 1.035 & 1.375 & 0.307 \\
    DWA   & 0.915 & 1.419 & 0.309 & 1.165 & 2.232 & 0.248 & \underline{0.899} & 1.005 & 0.442 & 1.040 & 1.372 & 0.312 \\
    MGDA  & 0.845 & 1.328 & 0.292 & 1.075 & 2.058 & 0.237 & 0.809 & 1.104 & 0.439 & \underline{1.104} & \underline{1.673} & \underline{0.350} \\
    MTAdam & 0.869 & 1.382 & 0.305 & 1.112 & 2.153 & 0.247 & 0.880 & \underline{1.015} & \underline{0.463} & 1.071 & 1.525 & 0.348 \\
	\midrule
    GradSimilarity & 0.923 & 1.444 & 0.308 & 1.186 & 2.270 & 0.255 & 0.817 & 0.977 & 0.427 & 1.025 & 1.529 & 0.336 \\
    GradSurgery & \underline{0.936} & \underline{1.471} & 0.319 & \underline{1.213} & \underline{2.371} & \underline{0.263} & 0.876 & 0.998 & 0.445 & 1.042 & 1.434 & 0.327 \\
    OL-AUX & 0.931 & \underline{1.471} & 0.311 & 1.162 & 2.224 & 0.243 & 0.804 & 0.921 & 0.413 & 0.950 & 1.312 & 0.295 \\
	\midrule
	MetaBalance & \textbf{0.990$^{*}$} & \textbf{1.550$^{*}$} & \textbf{0.339$^{*}$} & \textbf{1.258$^{*}$} & \textbf{2.421$^{*}$} & \textbf{0.269$^{*}$} & \textbf{0.974$^{*}$} & \textbf{1.164$^{*}$} & \textbf{0.509$^{*}$} & \textbf{1.134$^{*}$} & \textbf{1.588} & \textbf{0.353} \\
	\midrule
    Improvement & 5.77\%  & 5.32\%  & 3.96\%  & 3.66\%  & 2.09\%  & 2.08\%  & 8.34\%  & 14.68\% & 10.01\% & 2.72\%  & --  & 0.86\% \\
	\bottomrule
    \end{tabular}%
	\\
    {\raggedright  $*$ We conduct a two-sided significant test between MetaBalance and the strongest baseline (highlighted by underscore), where * means the p-value is smaller than 0.05. \par}
  \label{tab:Experimental Results}%
\end{table*}%

\subsection{RQ2: Comparison with Baseline Methods}
\label{sec: Experimental Results and Analysis}
Table \ref{tab:Experimental Results} presents the experimental results and the improvement of MetaBalance upon the strongest baseline in terms of each metric, where MetaBalance significantly outperforms all baselines over most of metrics on the two datasets.

\begin{figure}[]
    \centering
    \setlength{\abovecaptionskip}{0.0cm}
  \setlength{\belowcaptionskip}{-0.5cm}
    \includegraphics[width=.6\linewidth]{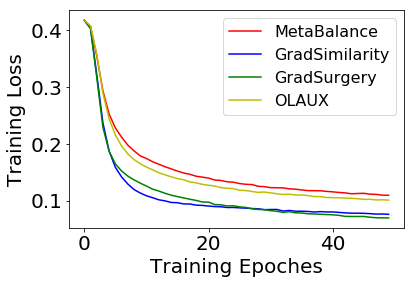}
    \caption{The training loss on UserBehavior-2017}
    \label{Fig:The training loss on UserBehavior-2017 dataset}
\end{figure}

\textbf{MetaBalance vs. gradient direction-based methods.} First, we observe that MetaBalance outperforms GradSimilarity, OL-AUX and GradSurgery, which are designed to boost the target task via adapting auxiliary tasks. Remember that the same idea behind these methods is that the less similar the direction of target gradient and one auxiliary gradient is, the lower weight will be assigned to that auxiliary task. While these gradient direction-based methods have worse performance than MetaBalance over the testing dataset, interestingly, they actually achieve better \textbf{training} loss than MetaBalance, where an example is shown in Figure \ref{Fig:The training loss on UserBehavior-2017 dataset}, which demonstrates they are more prone to overfitting than MetaBalance. Hence, this observation reveals that auxiliary gradients that have dissimilar directions with the target gradient might be sometimes helpful to improve the generalization ability of the model, which is consistent with the observations in the literature \cite{vandenhende2020multi, chen2020just}. For example, they might help the target task to correct its direction of the optimization to achieve a better generalization ability. MetaBalance keeps the direction conflicts between the target gradient and the auxiliary gradients but reduces the auxiliary gradients whose magnitudes are much larger than the target gradient, which prevents the dominance of auxiliary tasks and shows more robust performance for personalized recommendations. 

In addition, we are curious if MetaBalance can be enhanced when it also considers using the direction similarity to adapt auxiliary gradients. Specifically, in each training iteration, we first enlarge or reduce auxiliary gradients via MetaBalance and then enlarge or reduce them again according to one of the gradient direction-based methods. The results in Table \ref{tab:MetaBalance plus Gradient Direction-based Methods} show that the performance of MetaBalance mostly drops after including the gradient direction-based methods, which demonstrates that naively combining both magnitude and direction-based approaches can interfere with each another. We leave how to better consider both gradient magnitudes and directions for adapting auxiliary tasks to help the target task in the future work. 



\textbf{MetaBalance vs. multi-task balancing methods.} Second, it is understandable that Uncertainty, GradNorm, DWA are inferior to MetaBalance because they have no special preference to the target task. In DWA, the lower the loss decreases, the higher the weight is assigned to that loss. In GradNorm, the target task gradient magnitude is regularized to be similar to the average of all gradient magnitudes, which might not be the optimal magnitude for the target task optimization. In Uncertainty, the higher the uncertainty of the task dataset, the higher weight is assigned to that task loss. We also compare MGDA \cite{sener2018multi} as one of the most representative Pareto methods with MetaBalance. MGDA treats multi-task learning as multi-objective optimization problem and finds solutions that satisfy Pareto optimality. In MGDA, the shared parameters are only updated along common directions of the gradients for all tasks, which might not be the best optimization direction for the target task. Consequently, the target task is not guaranteed to be improved the most among all tasks in Pareto optimal solutions like MGDA. In contrast, MetaBalance is a specialized method designed for boosting the target task. As Table \ref{tab:Experimental Results} shows, MetaBalance significantly outperforms MGDA over most of the metrics. Although MTAdam is not originally designed for auxiliary learning, we let the target task serve as the anchor task in MTAdam. In this way, MetaBalance and MTAdam share the same core idea that the auxiliary gradient magnitudes become closer to the target gradient. However, Table \ref{tab:Experimental Results} shows that MetaBalance significantly outperforms MTAdam. The possible reason might be the relax factor in MetaBalance that can control the magnitude proximity, which makes MetaBalance more flexible than MTAdam. 

In addition, Vanilla-Multi is even inferior to Single-loss over most of metrics on both datasets. This demonstrates that transfer learning from auxiliary tasks is a non-trivial task -- that might hurt the performance of the target task rather than boosting it. After that, Table \ref{tab:Experimental Results} shows that Weights-Tuning, where the target task loss normally has a higher weight assigned than the auxiliary tasks, outperforms Vanilla-Multi over all metrics on both datasets. However, the performance of Weights-Tuning is significantly inferior to MetaBalance. A possible reason is that the tuned weights are fixed during the training and hence behave sub-optimally in adapting auxiliary tasks. 

To sum up, the results demonstrate that the three strategies and the relax factor make MetaBalance a flexible and effective framework to adapt auxiliary tasks from the perspective of gradient magnitudes, which significantly improves the target task's performance and outperforms baselines. 


\begin{table}[htbp]
  \centering
     \setlength{\abovecaptionskip}{0.0cm}
  \setlength{\belowcaptionskip}{0.0cm}
\small
\renewcommand\arraystretch{1.0}
\setlength{\tabcolsep}{2.0pt}
  \caption{MetaBalance plus Gradient Direction-based Methods}

    \begin{tabular}{lcccccc}
\toprule
    Metric(\%) & N@10  & R@10  & P@10  & N@20  & R@20  & P@20 \\
\midrule
    MetaBalance (MB) & 0.990 & 1.550 & 0.339 & 1.258 & 2.421 & 0.269 \\
    MB+GradientSimilarity & 0.937 & 1.398 & 0.311 & 1.190 & 2.210 & 0.250 \\
    MB+GradientSurgery & 0.925 & 1.585 & 0.329 & 1.167 & 2.351 & 0.258 \\
    MB+OL-AUX & 0.898 & 1.374 & 0.308 & 1.158 & 2.224 & 0.248 \\
\bottomrule
    \end{tabular}%
  \label{tab:MetaBalance plus Gradient Direction-based Methods}%
\end{table}%

\subsection{RQ3: Collaboration with More Optimizers}
\label{appen: Collaboration with More Optimizers} 
\begin{figure}
  \centering
  \setlength{\abovecaptionskip}{0.0cm}
 \setlength{\belowcaptionskip}{-0.2cm}
 \subfigure[Adagrad]{
    \label{visual: MetaBalance with Adagrad} 
    \includegraphics[width=1.6in]{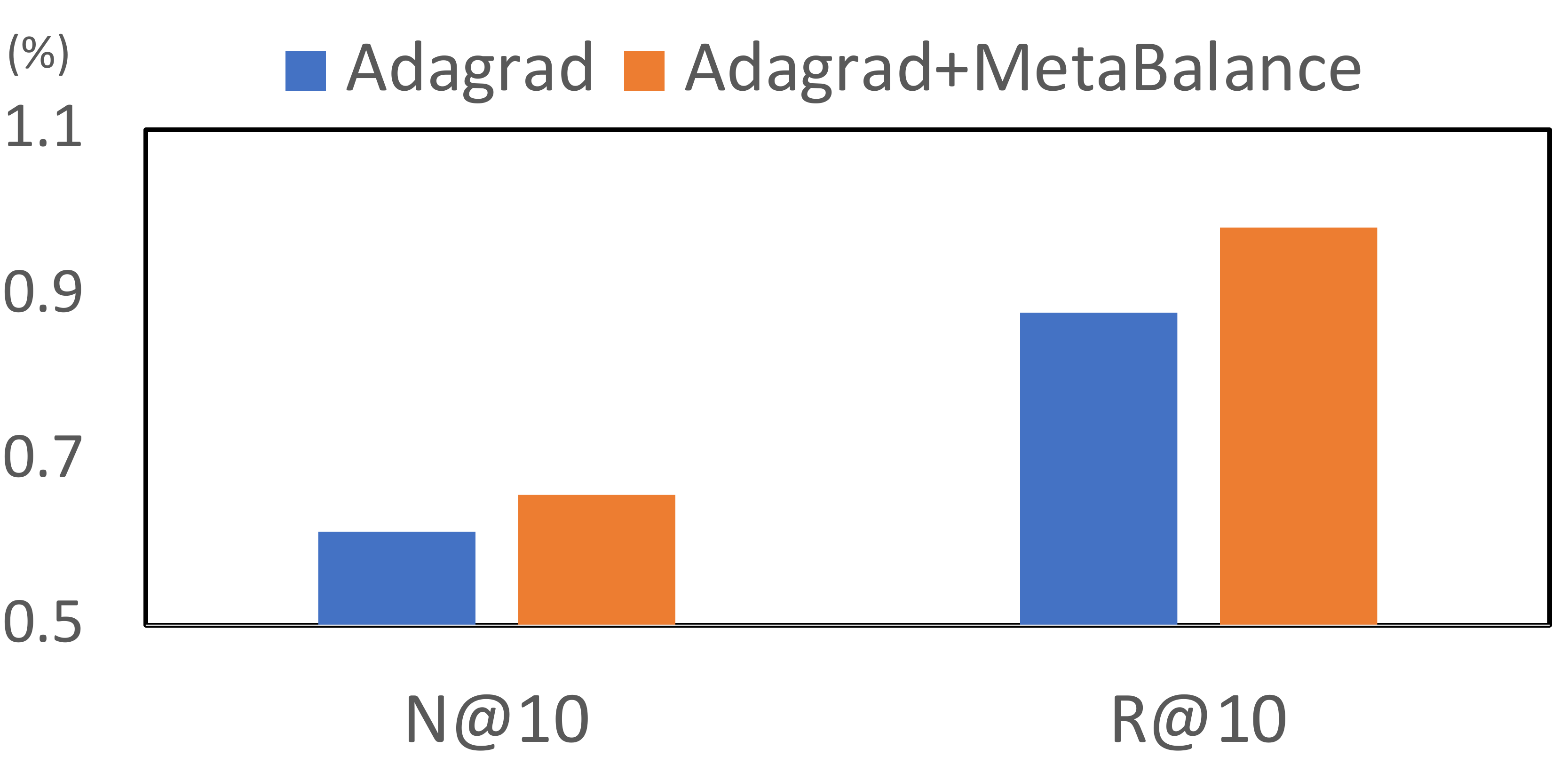}}
  \subfigure[RMSProp]{
    \label{visual: MetaBalance with RMSProp} 
    \includegraphics[width=1.6in]{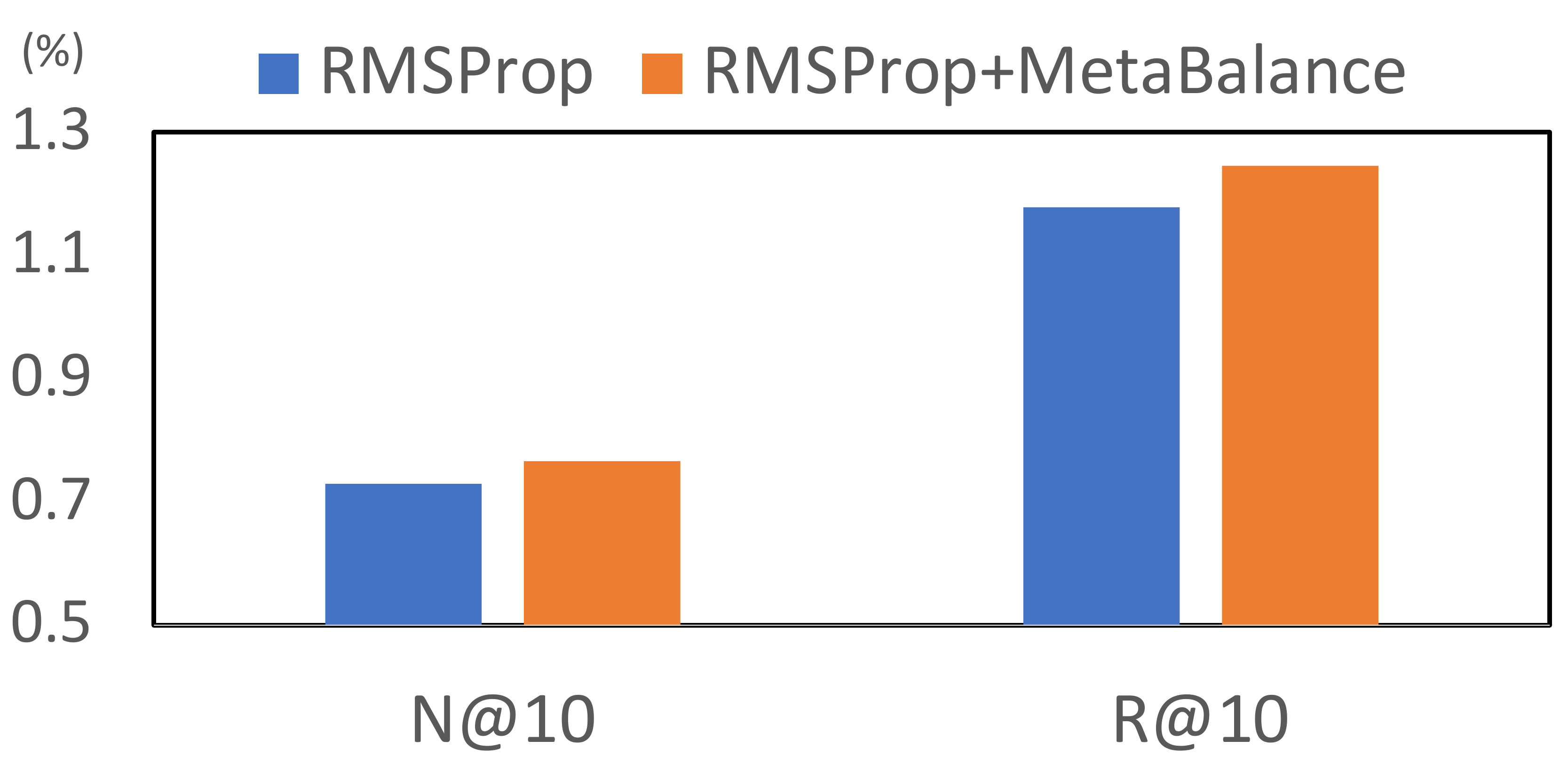}}
   \caption{Collaboration with Other Optimizers}
  \label{fig: collaborate with optimizers} 
\end{figure}

As shown in Algorithm \ref{alg:Basic MetaBalance} and \ref{alg:The Complete Version of MetaBalance}, MetaBalance balances the gradient magnitudes and these balanced gradients are used to update shared parameters following the rules of optimizers. Results in Table \ref{tab:Experimental Results} have shown that MetaBalance can collaborate with Adam well. We are also curious if MetaBalance can collaborate with other popular optimizers -- achieving higher performance for the target task compared to the multi-task network that is trained without MetaBalance. In Figure \ref{fig: collaborate with optimizers}, we observe that two other widely used optimizers -- Adagrad \citep{duchi2011adaptive} and RMSProp \citep{Tieleman2012} -- can also achieve better performance via using the balanced gradients from MetaBalance. This result demonstrates that MetaBalance can flexibly collaborate with commonly-used optimizers.

\subsection{RQ4: Impact of Moving Averages of Gradient Magnitudes}
\label{appen: Ablation Study of Moving Average of Magnitude} 
In Table \ref{tab:Ablation Study of Moving Average of Magnitude}, we compare the performance of MetaBalance with its variant (``$-$Moving Average'') where the moving averages of magnitude $m_{tar}^{t}$ and $m_{aux, i}^{t}$ (in Equation \ref{equation: moving average} and \ref{equation: moving average auxiliary task}) are replaced with the current magnitudes $\mathbf{G}_{tar}^{t}$ and $\mathbf{G}_{aux, i}^{t}$ at each iteration. We observe that the performance drops slightly on UserBehavior-2017 and drastically on IJCAI-2015 dataset. This result demonstrates the moving averages of magnitudes benefits the optimization, which takes into account the variance among all gradient magnitudes over the training iterations.
 
\begin{table}[htbp]
  \centering
  \setlength{\abovecaptionskip}{0.0cm}
  \setlength{\belowcaptionskip}{0.0cm}
\small
  \caption{Ablation Study of Moving Average of Magnitude}
    \begin{tabular}{lccc|ccc}
\toprule
    Datasets      & \multicolumn{3}{c|}{UserBehavior-2017} & \multicolumn{3}{c}{IJCAI-2015} \\
\midrule
    Metrics (\%)     & N@10  & R@10  & P@10  & N@10  & R@10  & P@10 \\
\midrule
    MetaBalance & 0.990 & 1.550 & 0.339 & 0.974 & 1.164 & 0.509 \\
    $-$Moving Average & 0.983 & 1.513 & 0.325 & 0.835 & 0.956 & 0.426 \\
\bottomrule
    \end{tabular}%
  \label{tab:Ablation Study of Moving Average of Magnitude}%
\end{table}%

\section{Conclusion}
In many personalized recommendation scenarios, the target task can be improved via training auxiliary tasks alongside this target task on a multi-task network. In this paper, we propose MetaBalance to adapt auxiliary tasks to better assist the target task from the perspective of gradient magnitude. Specifically, MetaBalance has three adapting strategies, such that it not only protects the target task from the dominance of auxiliary tasks but also avoids that one or more auxiliary tasks are ignored. Moreover, auxiliary gradients are balanced dynamically throughout the training and adaptively for each part of the network. Our experiments show that MetaBalance can be flexibly adapted to different scenarios and significantly outperforms previous methods on two real-world datasets.

\clearpage

\bibliographystyle{ACM-Reference-Format}
\bibliography{sample-bibliography}

\clearpage

\appendix

\section{Appendix}

\subsection{Relations of MetaBalance to Previous Methods}
In this section, we compare MetaBalance with previous methods.

\subsubsection{Auxiliary Task Adapting Methods} 
\label{sec: Adapting Auxiliary Tasks Methods}

These methods are specifically designed for auxiliary learning such that auxiliary tasks are adapted to better improve the target task. The common idea is that if $\mathbf{G}_{aux,i}$ is conflicting with $\mathbf{G}_{tar}$ from the perspective of direction, $\mathbf{G}_{aux,i}$ will be down-weighted or masked out. Compared with them, MetaBalance is the first method that adapts auxiliary task to assist the target task from the perspective of gradient magnitudes rather than punishing $\mathbf{G}_{aux,i}$ due to conflicting directions. The experimental results show that keeping inner-competition between the target gradient and conflicting auxiliary gradients as MetaBalance does improves the generalization ability of the model.

\smallskip
\noindent\textbf{GradSimilarity} \citep{du2018adapting} adapts auxiliary tasks via the gradient similarity between $\mathbf{G}_{tar}$ and $\mathbf{G}_{aux,i}$. Specifically, if $cosine(\mathbf{G}_{tar}, \mathbf{G}_{aux,i})$ is negative, then $\mathbf{G}_{aux,i}$ will not be added to $\mathbf{G}_{total}$ and hence will be ignored in updating the shared layers.

\smallskip
\noindent\textbf{GradSurgery\footnote{GradSurgery is originally for balancing multi-task learning. We can easily apply GradSurgery for auxiliary learning by specifying a task as the target task and the others as the auxiliary tasks}} \citep{yu2020gradient} replaces $\mathbf{G}_{aux,i}$ by its projection onto the normal plane of $\mathbf{G}_{tar}$ if $cosine(\mathbf{G}_{tar}, \mathbf{G}_{aux,i})$ is negative, unlike \citep{du2018adapting} where $\mathbf{G}_{aux,i}$ is just ignored. Formally, if $cosine(\mathbf{G}_{tar}, \mathbf{G}_{aux,i})$ is negative, they let:
\begin{equation}
	\mathbf{G}_{aux,i}^{t} = \mathbf{G}_{aux,i}^{t} - \frac{\mathbf{G}_{aux,i}^{t} \cdot \mathbf{G}_{tar}^{t}}{\| \mathbf{G}_{tar}^{t} \|^{2}} \cdot \mathbf{G}_{tar}^{t}
\end{equation}
In this way, the conflict between $\mathbf{G}_{tar}$ and $\mathbf{G}_{aux,i}$ can be alleviated. 

\smallskip
\noindent\textbf{OL-AUX} (Oline learning for Auxiliary Losses) \citep{lin2019adaptive} defines the total loss as $\mathcal{L}^{t} = \mathcal{L}_{tar}^{t} + \sum_{i=0}^{K}w_{aux,i}^{t} \cdot \mathcal{L}_{aux, i}^{t}$ where $w_{aux,i}$ is the weight of $\mathcal{L}_{aux, i}$ and $\mathcal{V}^{t}(\mathbf{w})$ as the speed at which the target task loss decreases at the $t$-th iteration and $\mathbf{w}=[w_{1},...,w_{K}]^{T}$. OL-AUX seek to optimize the N-step decrease of the target task w.r.t $\mathbf{w}$:
\begin{equation}
	\mathcal{V}^{t,t+N}(\mathbf{w}) = \mathcal{L}_{tar}^{t+N} - \mathcal{L}_{tar}^{t} 
\end{equation}

With some approximations, they find that $\forall i=1,...,K$:

\begin{equation}
	\nabla_{w_{aux,i}}\mathcal{V}^{t,t+N}(w_{aux,i}) = -\sum_{j=0}^{N-1} (\mathbf{G}_{tar}^{t+j})^{T} \mathbf{G}_{aux, i}^{t+j}
\end{equation}
Then, $\mathbf{w} \leftarrow \mathbf{w} - \beta \cdot \nabla_{\mathbf{w}}\mathcal{V}^{t,t+N}(\mathbf{w})$ such that the speed at which $\mathcal{L}_{tar}$ decreases could be maximized. 

\subsubsection{Multi-Task Balancing Methods}
\label{sec:Multi-Task Balancing Methods}
In contrast to the auxiliary learning-specific methods, these  multi-task balancing methods are for general learning where all tasks are treated equally important. Although they have no preference to the target task, they are valid baselines because MetaBalance is specifically for auxiliary learning and is supposed to outperform them. For convenience, we let $j$ denotes the index of any task in this subsection, where $j = 0,1,...,K$. Let $w_{j}$ be the weight of the $j$-th task loss $\mathcal{L}_{j}$.

\smallskip
\noindent\textbf{Uncertainty}  \citep{kendall2018multi} assumes that the higher the uncertainty of task data is, the lower the weight of this task loss should be assigned. They design a learnable parameter $\sigma_{j}$ to model the uncertainty for each task. Specifically, they optimize the model parameters and $\sigma_{j}$ to minimize the following objective:
\begin{equation}
	\mathcal{L} = \sum_{j=0}^{K} \frac{1}{\sigma_{j}^{2}}\mathcal{L}_{j} + \sum_{j=0}^{K}log\sigma_{j}
\end{equation}
Minimizing the loss $\mathcal{L}$ w.r.t. $\sigma_{j}$ can automatically balance $\mathcal{L}_{j}$ during training, where increasing $\sigma_{j}$ reduces the weight for task loss $\mathcal{L}_{j}$. 

\smallskip
\noindent\textbf{GradNorm} \citep{chen2018gradnorm} encourages $\| \mathbf{G}_{j}^{t} \|$ to be the mean of all $\| \mathbf{G}_{j}^{t} \|, \ j=0,...,K$. In this way, all tasks could have a similar impact on the updating of shared-parameters. In particular, they minimize the following two objectives:
\begin{gather}
	\mathcal{L}^{t} = \sum_{j=0}^{K} w_{j}^{t} \cdot \mathcal{L}_{j}^{t} \\
	\mathcal{L}_{normLoss}^{t} = \sum_{j=0}^{K} L1Norm(\| w_{j}^{t} \cdot \mathbf{G}_{j}^{t} \| - \| \overline{\mathbf{G}^{t}} \| \cdot [r_{j}^{t}]^{\alpha})
\end{gather}
	
In each iteration, $\mathcal{L}^{t}$ is firstly optimized w.r.t model parameters $\theta$ (not including $w_{j}^{t}$) to obtain $\mathbf{G}_{j}^{t}$ and the $\mathcal{L}_{normLoss}^{t}$ is optimized w.r.t $w_{j}^{t}$. In the next iteration, updated $w_{j}^{t}$ can balance $\mathcal{L}_{j}^{t}$. Moreover, $r_{j}^{t}$ is to model the pace at which different tasks are learned, where $r_{j}^{t} = p_{j}^{t} /  E[p_{j}^{t}]$ and $p_{j}^{t} = \mathcal{L}_{j}^{t} / \mathcal{L}_{j}^{0}$. And $\alpha$ is a hyper-parameter which sets the strength of forcing tasks back to a common training rate. 

\smallskip
\noindent\textbf{DWA} (Dynamic Weight Averaging) \citep{liu2019end} balances the pace at which tasks are learned. In DWA, $w_{j}^{t}$ is set as:
\begin{equation}
	w_{j}^{t} = \frac{N \cdot exp(p_{j}^{t-1}/T)}{\sum_{n} exp(p_{n}^{t-1}/T)}, \ \ p_{j}^{t-1} = \frac{\mathcal{L}_{j}^{t-1}}{\mathcal{L}_{j}^{t-2}}
\end{equation}

where $N$ is the number of tasks and temperature $T$ controls the softness of the task weighting in the softmax function. $p_{j}$ estimates the relative descending rate of $\mathcal{L}_{j}$. When $\mathcal{L}_{j}$ decreases at a slower rate compared with other task losses, $w_{j}$ will be increased. 

\smallskip
\noindent\textbf{MGDA} (Multiple-Gradient Descent Algorithm) \citep{sener2018multi} treats multi-task learning as multi-objective optimization problem and finds solutions that satisfies Pareto optimality  -- as long as there is a common direction along which losses can be decreased, we have not reached a Pareto optimal point yet. Since the shared parameters are only updated along \textbf{common} directions of the task-specific gradients, MGDA has no preference on a particular task.

\begin{table*}[htbp]
    \setlength{\abovecaptionskip}{0.0cm}
  \setlength{\belowcaptionskip}{0.0cm}
  \centering
\renewcommand\arraystretch{0.9}
\setlength{\tabcolsep}{2.4pt}
\small
  \caption{Statistics of Preprocessed Datasets}
    \begin{tabular}{lcc|cc|ccc}
	\toprule
          &       &       & \multicolumn{2}{c|}{Target Task} & \multicolumn{3}{c}{Auxiliary Tasks} \\
	\midrule
    Dataset & \multicolumn{1}{c}{\#User} & \multicolumn{1}{c|}{\#Item} & \multicolumn{1}{c}{\#Buy} & \multicolumn{1}{c|}{Density of Buy} & \multicolumn{1}{c}{\#Add-to-Cart} & \multicolumn{1}{c}{\#Click} & \multicolumn{1}{c}{\#Add-to-Favorite} \\
	\midrule
    IJCAI-2015 & 19,839 & 50,973 & 390,600 & 0.039\% & 1,693  & 2,025,910 & 224,279 \\
    UserBehavior-2017 & 16,089 & 25,813 & 89,404 & 0.022\% & 53,245 & 394,246 & 19,585 \\
	\bottomrule
    \end{tabular}%
	\\
\label{tab:Statistics of datasets}%
\end{table*}%

\smallskip
\noindent\textbf{MTAdam} \citep{malkiel2020mtadam} is an Adam-based optimizer that balances gradient magnitudes and then update parameters according to the rule of Adam \citep{kingma2014adam}. Following MTAdam, we also directly manipulate the gradient magnitudes, instead of weighting task losses like Uncertainty \citep{kendall2018multi}, GradNorm \citep{chen2018gradnorm} and DWA \citep{liu2019end}. MetaBalance differs from MTAdam in the following aspects:
\begin{itemize}
	\item MTAdam lets all gradient magnitudes be similar to that of the first loss (not necessarily the target loss) while MetaBalance has three strategies that can flexibly encourage auxiliary gradients to better help the target task optimization.
	\item $\| \mathbf{G}_{aux,i}^{t} \|$ can only be very similar to $\| \mathbf{G}_{tar}^{t} \|$ in MTAdam while MetaBalance can adjust the proximity of $\| \mathbf{G}_{aux,i}^{t} \|$ to $\| \mathbf{G}_{tar}^{t} \|$ via the relax factor, which is vital for adapting MetaBalance to different scenarios.
	\item MetaBalance is an auxiliary task adapting algorithm that can collaborate with most optimizers like Adagrad or RMSprop to update parameters, whereas MTAdam is specially designed for Adam-based optimizers only.
\end{itemize}

\subsection{Reproducibility of Experiments}
\label{appen: Reproducibility}
\textit{The code of our approach can be found at here.}\footnote{\url{https://github.com/facebookresearch/MetaBalance}}

\medskip
\noindent\textbf{Dataset Preprocessed and Split.} IJCAI-2015 is preprocessed by filtering out users who purchase fewer than 20 unique items and items which are purchased by fewer than 10 unique users. We omit add-to-cart as an auxiliary task in IJCAI-2015 because this behavior only has 1,693 feedbacks. For UserBehavior-2017, we filter out users who purchase fewer than 10 unique items and items which are purchased by fewer than 10 unique users. The  datasets are summarized in Table \ref{tab:Statistics of datasets}. We randomly split purchase interactions into a training set (70\%), validation set (10\%) and testing set (20\%). For the interactions of auxiliary tasks like add-to-cart, we merge them into the training set. Since auxiliary interactions like add-to-cart are highly related to purchase interaction, to prevent possible information leakage, we remove user-item pairs from the auxiliary interactions if these pairs appear in the validation set and testing set of the purchase interactions.

\medskip
\noindent\textbf{Implementation and Training Details.} We implement MetaBalance, Uncertainty, DWA, GradSimilarity, GradSurgery and OL-AUX via Pytorch. The code of GradNorm is from this repo\footnote{\url{https://github.com/hosseinshn/GradNorm}} and the code of MTAdam is from the authors.\footnote{\url{https://github.com/ItzikMalkiel/MTAdam}} All experiments are conducted on an Nvidia GeForce GTX Titan X GPU with 12 GB memory. Cross-entropy loss is adopted for each task and Adam \citep{kingma2014adam} is the optimizer with batch size of 256 and learning rate of 0.001. 

\medskip
\noindent\textbf{Hyper-parameters.} All hyper-parameters are carefully tuned in the validation set, where early stopping strategy is applied such that we terminate training if validation performance does not improve over 20 epochs. In the multi-task recommendation network, the size of user and item embeddings is 64, the size of the shared MLP layers is \{32, 16, 8\} and the size of the task-specific MLP layers is \{64, 32\}. To prevent overfitting, dropout with rate of 0.5 is applied for each layer and we also use weight decay with rate of e-7. For MetaBalance, $r$ is selected from 0.1, 0.2, ...0.9 and 0.7 is the best for UserBehavior-2017 and 0.9 is the best for IJCAI-2015. For MTAdam, $\beta_{1}$, $\beta_{2}$, $\beta_{3}$ are respectively set as 0.9, 0.999 and 0.9. For DWA, T is set as 2 and we calculate the mean of losses in very 5 iterations on IJCAI-2015 and in very 10 iterations on UserBehavior-2017. For GradNorm, $\alpha$ is set as 0.75 on IJCAI-2015 and 0 on UserBehavior-2017. For OL-AUX, $\beta$ is set as 0.1 on IJCAI-2015 and 1 on UserBehavior-2017.

\begin{figure}[]
    \centering
    \includegraphics[scale=0.19]{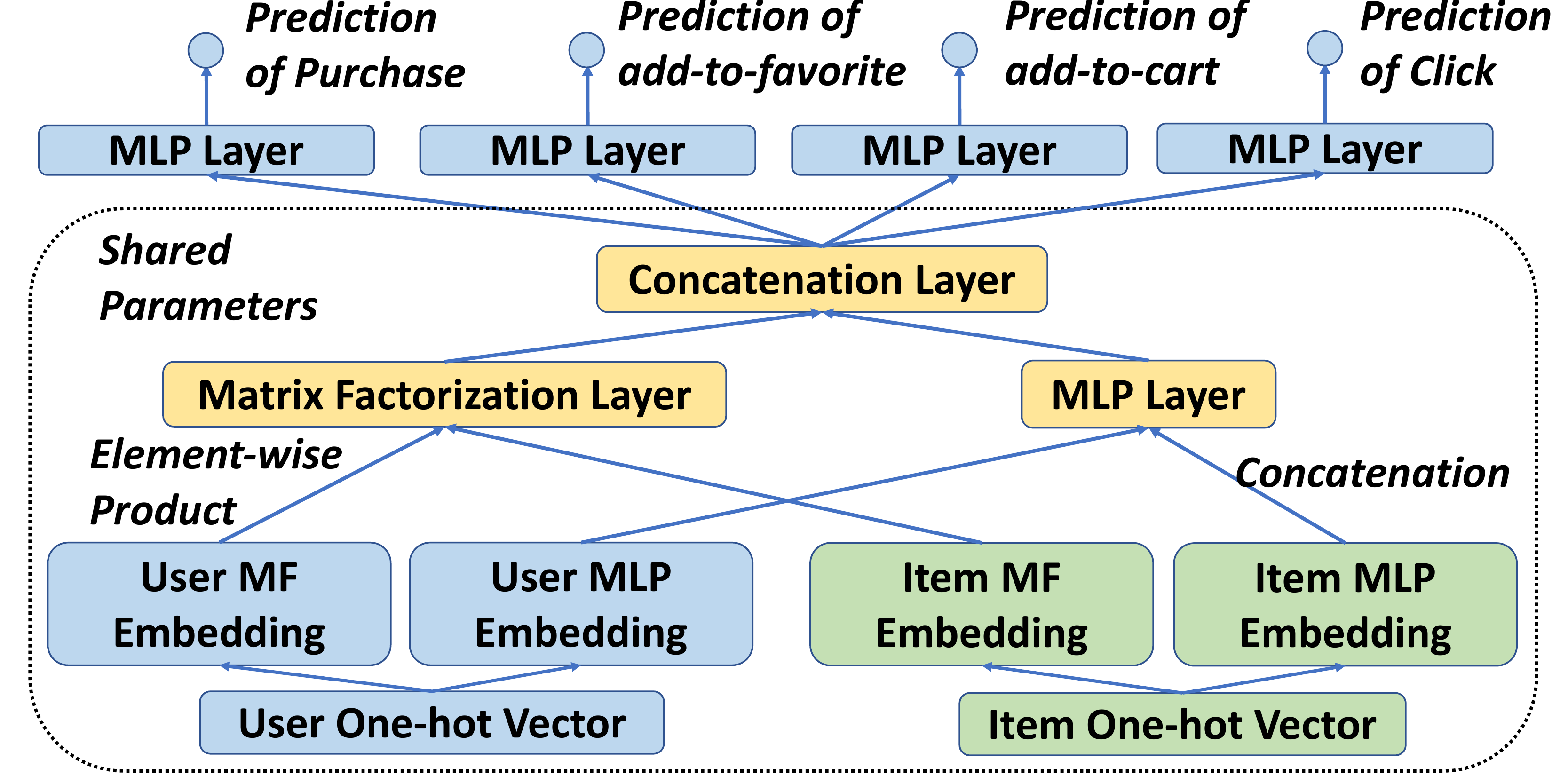}
    \caption{Multi-Task recommendation network in the evaluation. The shared bottom  layers is the combination of MLP layer and matrix factorization layer, which is widely adopted for recsys in both academia \citep{ncf} and industry \cite{cheng2016wide, naumov2019deep}}
    \label{fig: Multi-Task Recommendation Network}
\end{figure}

\end{document}